\documentclass[runningheads]{llncs}

\usepackage{graphicx}
\usepackage{balance}
\usepackage{caption}
\usepackage{makecell}
\usepackage{mdwtab}
\usepackage{subcaption}
\usepackage{soul}
\usepackage{tabularx}
\usepackage{listings}
\usepackage{color}
\usepackage{enumitem}
\usepackage{ulem}
\usepackage{multirow} 
\usepackage{url}
\usepackage{amsmath}

\begin{document}

\title{DeepCapture: Image Spam Detection Using Deep Learning and Data Augmentation}

\titlerunning{DeepCapture}

\author{Bedeuro Kim\inst{1,2}\and
Sharif Abuadbba\inst{2,3}\and
Hyoungshick Kim\inst{1,2}}

\institute{Sungkyunkwan University, Suwon, Republic of Korea\\ \email{\{kimbdr,hyoung\}@skku.edu}\\
\and
Data61, CSIRO, Sydney, Australia\\
\and
Cyber Security Cooperative Research Centre, Australia\\
\email{\{Bedeuro.Kim,Sharif.Abuadbba,Hyoung.Kim\}@data61.csiro.au}
}

\maketitle         
\begin{abstract}
Image spam emails are often used to evade text-based spam filters that detect spam emails with their frequently used keywords. In this paper, we propose a new image spam email detection tool called DeepCapture using a convolutional neural network (CNN) model. There have been many efforts to detect image spam emails, but there is a significant performance degrade against entirely new and unseen image spam emails due to overfitting during the training phase. To address this challenging issue, we mainly focus on developing a more robust model to address the overfitting problem. Our key idea is to build a CNN-XGBoost framework consisting of eight layers only with a large number of training samples using data augmentation techniques tailored towards the image spam detection task. To show the feasibility of DeepCapture, we evaluate its performance with publicly available datasets consisting of 6,000 spam and 2,313 non-spam image samples. The experimental results show that DeepCapture is capable of achieving an F1-score of 88\%, which has a 6\% improvement over the best existing spam detection model CNN-SVM~\cite{shang2016image} with an F1-score of 82\%. Moreover, DeepCapture outperformed existing image spam detection solutions against new and unseen image datasets.

\end{abstract}

\section{Introduction}
\label{sec:intro}

Image-based spam emails (also referred to as ``image spam emails'') are designed to evade traditional text-based spam detection methods by replacing sentences or words contained in a spam email with images for expressing the same meaning~\cite{ismail2019image}. As image spam emails become popular~\cite{emailadvertisement}, several spam detection methods~\cite{Fumera06:spam,kim2017analysis,kumar2017svm} have been proposed to detect image spam emails with statistical properties of image spam emails (e.g., the ratio of text contents in an email sample). However, these countermeasures have disadvantages due to a high processing cost for text recognition in images~\cite{attar2013survey}. Recently, a convolutional neural network (CNN) model-based detection~\cite{shang2016image} was presented to address this processing cost issue and improve the detection accuracy. The recent advance of deep learning technologies in the image domain would bring a new angle or approach to security applications. CNN has the potential to process raw data inputs (e.g., the input image itself) by extracting important (low-level) features in an automated manner~\cite{Krizhevsky17:dnn}. However, we found that the detection accuracy of the existing CNN based image spam detection model~\cite{shang2016image} could be degraded significantly against new and unseen image spam emails.

To overcome the limitation of existing image spam detectors against new and unseen datasets, we propose a new image spam email detection tool called DeepCapture. DeepCapture consists of two phases: (1) data augmentation to introduce new training samples and (2) classification using a CNN-XGBoost model. In this paper, we focus on developing new data augmentation techniques tailored for image spam training dataset and designing an effective CNN architecture capable of detecting images used for spam emails with the optimized configuration for number of layers, number of filters, filter size, activation function, a number of epochs and batch size. 



To examine the feasibility of DeepCapture, we evaluate the performance of DeepCapture compared with existing image spam email detectors such as RSVM based detector~\cite{annadatha2018image} and CNN-SVM based detector~\cite{shang2016image}. In our experiments, we use a dataset consisting of 6,000 spam and 2,313 non-spam (hereinafter referred to as ham) image samples collected from real-world user emails. We also use our data augmentation techniques to balance the distribution of ham and spam samples and avoid performance degradation against new and unseen datasets. We evaluate the performance of the DeepCapture in two ways. First, we evaluate the performance of DeepCapture with/without data augmentation. Second, we also evaluate the performance of DeepCapture via cross data training scenarios with/without data augmentation. Our experimental results demonstrate that DeepCapture produced the best classification results in F1-score (88\%) compared with existing solutions. Moreover, for two cross data training scenarios against unseen datasets, DeepCapture also produced the best F1-score results compared with other classifiers. The use of data augmentation techniques would be necessary for processing new and unseen datasets. In the cross data training scenarios, F1-scores of all classifiers are less than 40\% without applying our data augmentation techniques.


This paper is constructed as follows: Section \ref{sec:Bg} describes the background of image spam email, convolutional neural network and data augmentation. Section \ref{sec:architecture} describes the model architecture of DeepCapture. Section \ref{sec:eva} describes experiment setups and evaluation results of DeepCapture. Section \ref{sec:Rw} describes the related work for image spam detection and we conclude in Section \ref{sec:CS}.

\section{Background}
\label{sec:Bg}
This section first presents the definition of the image spam email and then briefly provides the concept of a convolutional neural network. Finally, we present the description of data augmentation that is a widely used technique~\cite{Krizhevsky17:dnn} for improving the robustness of deep learning models. It increases the size of labeled training samples by leveraging task-specific data transformations that preserve class labels.

\subsection{Image Spam Email}

Since image spam emails appeared in 2004, several studies were conducted to formally define image spam emails and construct models to detect image spam emails in academia. Klangpraphan et al.~\cite{klangpraphant2010pimsi} observed that image spam emails contain an image-based link to a website, which looks like a text. Soranamageswari et al.~\cite{soranamageswari2010statistical} introduced the definition of image spam email as spam email having at least one image containing spam content.  

\begin{figure}[!ht]
\centering
\begin{subfigure}[b]{0.49\textwidth}{}
\centering
\includegraphics[width=\textwidth,height=4.3cm]{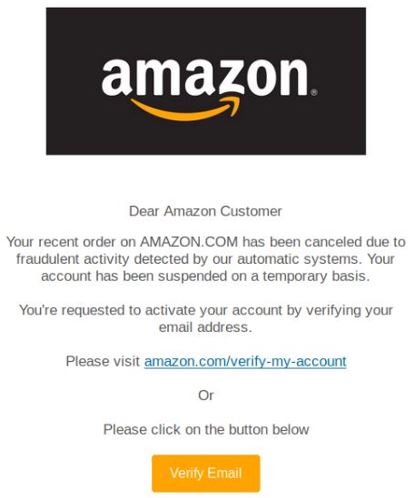}
\caption{Spam email containing a link}
\label{fig:y equals x}
\end{subfigure}
\hspace{0.15em}
\begin{subfigure}[b]{0.49\textwidth}
\centering
\includegraphics[width=\textwidth]{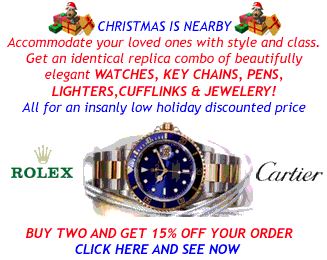}
\caption{Spam email showing an advertisement}
\label{fig:three sin x}
\end{subfigure}
\caption{Examples of image spam emails.}
\label{fig:imagespam}
\end{figure}

Figure \ref{fig:imagespam} shows two examples of image spam emails. In Figure \ref{fig:imagespam}(a), if a user clicks the ``Verify Email'' button, it tries to visit an attacker's website or download malware. In Figure \ref{fig:imagespam}(b), the spam image shows unwanted advertisement information to email recipients. Basically, the goal of image spam emails is to hide the attacker's message into an image for circumventing text-based spam filters. Based on this observation, in this paper, we define the image spam email as spam email with images displaying unwanted text information.

\subsection{Convolutional Neural Network (CNN)}

Convolutional neural network (CNN) is a kind of deep learning methods. Recently, in many classification tasks, CNN outperformed traditional machine learning methods. Therefore, it is widely believed that CNN has the potential to be used for security applications. 

CNN can automatically extract features of target objects from lower to higher levels by using convolutional and pooling layers. Convolutional layers play a role in extracting the features of the input. A convolutional layer consists of a set of filters and activation functions. A filter is a function to emphasize key features that are used to recognize target objects. The raw input data is converted into feature maps with filters, which becomes more clear after processing the activation functions. A pooling layer (or sub-sampling) reduces the number of features, which prevents overfitting caused by a high number of features and improve the learning rate. Finally, feature map layers are used as the input layer for the fully connected classifier. These are popularly applied to computer vision tasks such as object recognition~\cite{bappy2016cnn}.


\subsection{Data Augmentation}

In a classification problem, it is widely known that the performance of classifiers deteriorates when an imbalanced training dataset is used. If the number of instances in the major class is significantly greater than that in the minor class, the classification performance on the major class will be higher, and vice versa.

\begin{figure}[!ht]
\centering
\begin{subfigure}[t]{.235\linewidth}
\centering
\captionsetup{justification=centering}
\includegraphics[width=\textwidth]{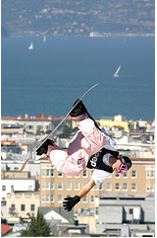}
\caption{Original image}
\label{fig:D_spam}
\end{subfigure}
\hspace{0.05em}
\begin{subfigure}[t]{.235\linewidth}
\centering
\captionsetup{justification=centering}
\includegraphics[width=\textwidth]{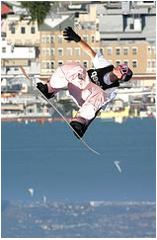}
\caption{Flipping}
\label{fig:D_spam2}
\end{subfigure}
\hspace{0.05em}
\begin{subfigure}[t]{.235\linewidth}
\centering
\includegraphics[width=\textwidth]{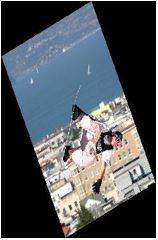}
\caption{Rotation}
\label{fig:D_ham}
\end{subfigure}
\hspace{0.05em}
\begin{subfigure}[t]{.235\linewidth}
\centering
\captionsetup{justification=centering}
\includegraphics[width=\textwidth]{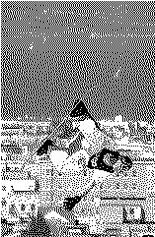}
\caption{Color transformation}
\label{fig:D_ham2}
\end{subfigure}
\caption{Examples of the image manipulation techniques.}
\label{fig:mani}
\end{figure}

Data augmentation is a popularly used method to solve the imbalance problem~\cite{shorten2019survey}, which increases the number of instances in minority classes to balance between majority classes and minority classes. In the image domain, new samples are typically generated by applying the geometric transformations or adding noise to training samples. Figure \ref{fig:mani} shows typically used image manipulation techniques such as flipping, rotation, and color transformation for image applications.

In the image spam detection problem, however, the effects of such general data augmentation techniques would be limited because the typical ham and spam images are different from the samples generated from such augmentation techniques. Therefore, in this paper, we focus on developing data augmentation for image ham and spam emails.


\section{Overview of DeepCapture}
\label{sec:architecture}

\begin{figure}[!ht]
\centering
\includegraphics[width=1\columnwidth]{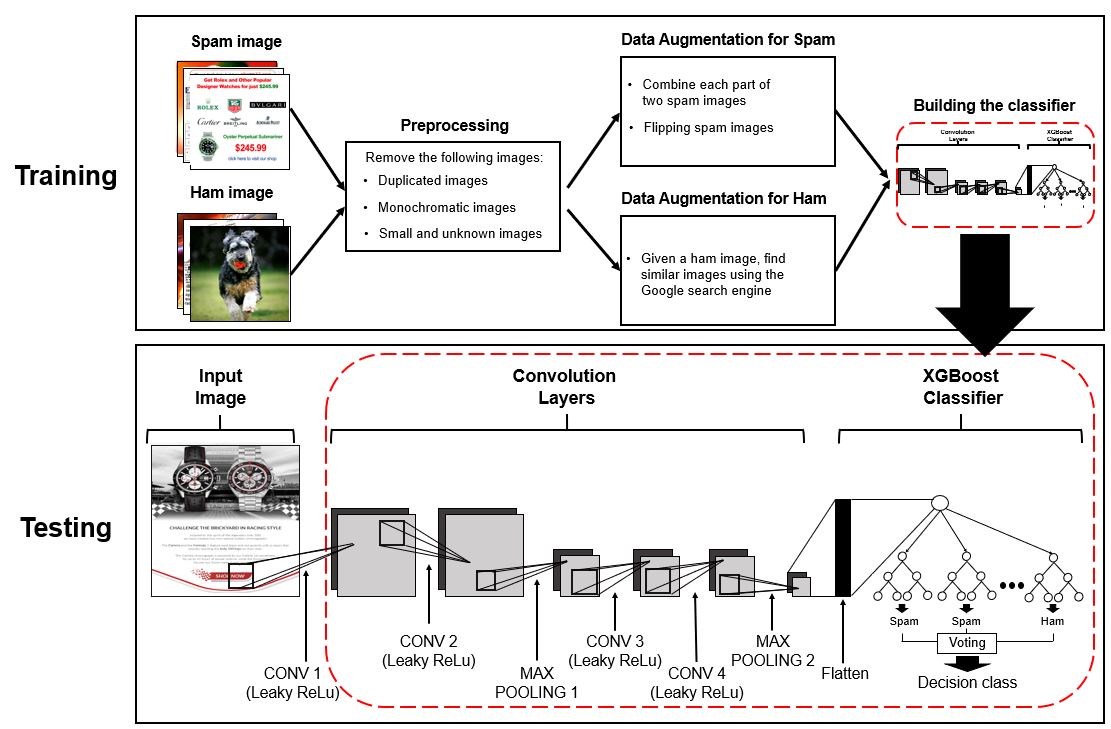}
\caption{Overview of DeepCapture.}
\label{fig:model}
\end{figure}

We designed DeepCapture using data augmentation and CNN to make it robust against new and unseen datasets. Figure \ref{fig:model} shows an overview of  DeepCapture architecture.

DeepCapture consists of two phases: (1) data augmentation to introduce new training samples and (2) classification using a CNN model.

\subsection{Data Augmentation in DeepCapture}
\label{sec:Data Augmentation in DeepCapture}

To address the class imbalance problem in image spam datasets and generalize the detection model, we introduce a new data augmentation method to create new ham and spam samples for training. The goal of data augmentation is to make augmented samples that are similar to real data.

For both ham and spam images, we commonly remove unnecessary images such as duplicate images, solid color background images, small and unknown images that cannot be recognized by human users. After removing unnecessary images, we apply different data augmentation methods to ham and spam images, respectively. 

\begin{figure}[!ht]
\centering
\includegraphics[width=1\columnwidth]{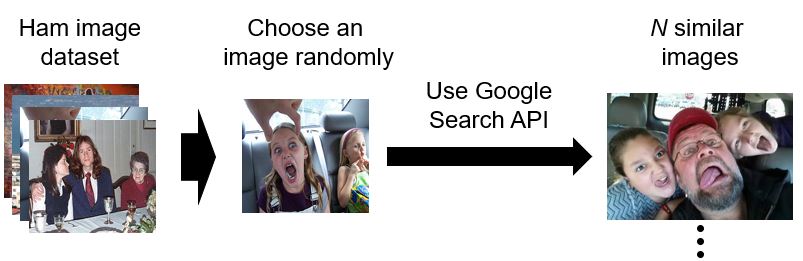}
\caption{Data augmentation process for ham images.}
\label{fig:similar}
\end{figure}

For ham images, we randomly choose an image among ham images and use an API to search images that are similar to the given image. For example, the Google Image Search API can be used to crawl the images similar to the ones we uploaded. For each uploaded image, $N$ (e.g., $N=100$) similar images can be obtained as ham-like images for training (see Figure~\ref{fig:similar}). Those images would be regarded as additional ham images because those are also actually used images on other websites.

\begin{figure}[!ht]
\centering
\includegraphics[width=1\columnwidth]{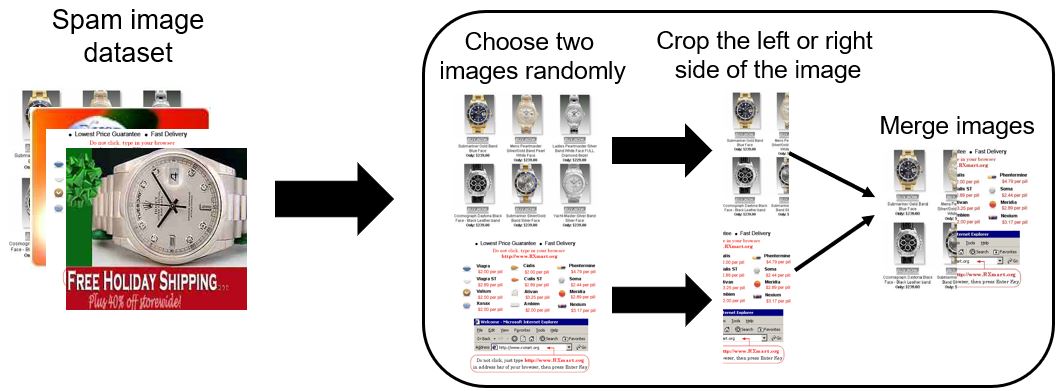}
\caption{Data augmentation process for spam images.}
\label{fig:cropflip}
\end{figure}

For spam images, we randomly choose two images among spam images and split each image in half from left to right (``left and right parts''). Next, we then combine the left part of an image with the right part of the other image. To combine parts from different images, we resize a part of an image so that its size is the same as the size of the part of another image (see Figure~\ref{fig:cropflip}). Our key observation is that a spam image typically consists of the image and text parts. Therefore, it is essential to create augmented samples having both image and text parts. Our data augmentation techniques are designed to produce such image samples.

\subsection{CNN-XGBoost Classification in DeepCapture}

As shown in Figure~\ref{fig:model}, the architecture of DeepCapture is composed of eight layers. Given an input image, the input image is resized to 32x32 pixels. The first six layers are convolutional layers, and the remaining two layers are used for the XGBoost classifier to determine whether a given image is spam or not. 

All convolutional layers use 3x3 kernel size and the Leaky ReLU function~\cite{maas2013rectifier}, which is used as the activation function. The Leaky ReLU function has the advantage to solve the gradient saturation problem and improve convergence speed. Unlike ReLU, in which the negative value is totally dropped, Leaky ReLU assigns a relatively small positive gradient for negative inputs. We also apply the 2x2 max pooling to the 3rd and 6th layers, which selects the maximum value from the prior feature map. The use of max pooling reduces the dimension of feature parameters and can help extract key features from the feature map. We also use regularization techniques to prevent the overfitting problem. We specifically use both L2 regularization~\cite{ng2004feature} and the dropout method~\cite{srivastava2014dropout} as regularization techniques for DeepCapture. 




After extracting features from the input image through the convolutional layers, we use the XGBoost~\cite{chen2016xgboost} classifier, which is a decision-tree-based ensemble machine learning algorithm that uses a gradient boosting framework. XGBoost builds a series of gradient boosted decision trees in a parallel manner and makes the final decision using a majority vote over those decision trees. In many situations, a CNN model typically uses the fully connected layers. However, for the image spam detection problem, we found that we can improve the detection accuracy if we replace the fully connected layers with a classifier such as XGBoost. We use the random search~\cite{bergstra2012random} method for optimizing hyperparameters used in the XGBoost classifier.



\section{Evaluation}
\label{sec:eva}
This section presents the performance evaluation results of DeepCapture (presented in Section~\ref{sec:architecture}) compared with state-of-the-art classification methods:  SVM~\cite{annadatha2018image}, RSVM~\cite{annadatha2018image} and CNN-SVM~\cite{shang2016image}.

\subsection{Dataset}

\begin{figure}[!ht]
\centering
\begin{subfigure}[t]{.235\linewidth}
\centering
\captionsetup{justification=centering}
\includegraphics[width=\textwidth]{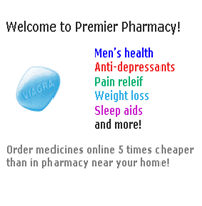}
\caption{Personal spam}
\label{fig:D_spam}
\end{subfigure}
\hspace{0.05em}
\begin{subfigure}[t]{.235\linewidth}
\centering
\captionsetup{justification=centering}
\includegraphics[width=\textwidth]{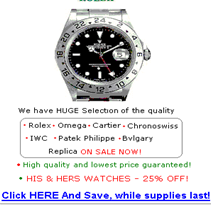}
\caption{SpamArchive spam}
\label{fig:D_spam2}
\end{subfigure}
\hspace{0.05em}
\begin{subfigure}[t]{.235\linewidth}
\centering
\includegraphics[width=\textwidth]{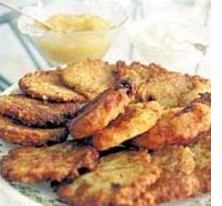}
\caption{Personal ham}
\label{fig:D_ham}
\end{subfigure}
\hspace{0.05em}
\begin{subfigure}[t]{.235\linewidth}
\centering
\captionsetup{justification=centering}
\includegraphics[width=\textwidth]{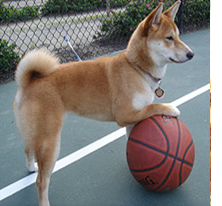}
\caption{Normal image ham}
\label{fig:D_ham2}
\end{subfigure}
\caption{Examples of image datasets.}
\label{fig:imagesample}
\end{figure}

To evaluate the performance of image spam email detection models, we use publicly available mixed datasets with two spam (``Personal spam'' and ``SpamArchive spam'') and two ham (``Personal ham'' and ``Normal image ham'') image datasets. Figure \ref{fig:imagesample} shows examples of those datasets.


In the ``Personal spam'' dataset, spam images were collected from 10 email accounts for one month, and ham images were collected from two email accounts for two years. The ``SpamArchive spam'' dataset~\cite{dredze2007learning} was constructed with many anonymous users. ``Normal image ham'' dataset~\cite{gao2008image} was collected from a photo-sharing website called ``Flickr'' (\url{https://www.flickr.com/}) and 20 scanned documents. From those datasets, we removed unnecessary image samples such as duplicated images, solid color background images, small and unknown images. In particular, since the ``SpamArchive spam'' dataset contains a lot of duplicated images such as the advertisement for a watch or corporate logo, we need to remove such duplicated images. After eliminating image samples that cannot be categorized as either normal ham or spam images, we were left with a dataset of 8,313 samples for experiments. The details of the dataset are presented in Table \ref{tab:dataset}. In the final dataset, the number of spam images is 6,000, while the number of ham images is 2,313. The ratio of ham to spam is around 1:3.

\begin{table}[ht!]
\centering
\caption{Description of the datasets.}
\label{tab:dataset}
\vspace{2mm}
\begin{tabular}{ |p{1.5cm}||p{3cm}|p{2cm}|}
\Xhline{3\arrayrulewidth}
\hline
 Category& Corpus &Total count\\
 \hline
 Spam   & Personal spam    & 786\\
        & SpamArchive spam  & 5,214\\
        \cline{2-3}
        & Total & 6,000\\
        \hline
 Ham    & Personal ham    & 1,503\\
        & Normal image ham  & 810 \\
        \cline{2-3}
        & Total & 2,313\\
\hline
\Xhline{3\arrayrulewidth}
\end{tabular}
\end{table}




\subsection{Experiment setup}

Our experiments were conducted using the Google Colab environment (\url{https://colab.research.google.com/}). It supports a GPU Nvidia Tesla K80 with 13GB of memory and an Intel(R) Xeon(R) CPU at 2.30GHz. We use Keras framework with the scikit-learn library in Python 3 to implement DeepCapture. 

For classification, we randomly divided 8,313 samples into a training set (60\%) and a testing set (40\%) with similar class distributions.

To address the data imbalance issue and make the classifier more robust against new and unseen image datasets, we use the data augmentation (DA) techniques presented in Section~\ref{sec:Data Augmentation in DeepCapture} to create additional image samples. Finally, we obtained 5,214 ham-like and 4,497 spam-like image samples with the images in the training set through our data augmentation techniques. Those image samples are used for training only.


\subsection{Classification results}

To evaluate the performance of classifiers, we use the four following metrics:

\begin{itemize}
\item \textbf{Accuracy (Acc.)}: the proportion of correctly classified images;
\item \textbf{Precision (Pre.)}: the proportion of images classified as spam that actually are spam;
\item \textbf{Recall (Rec.)}: the proportion of spam images that were accurately classified;
\item \textbf{F1-score (F1.)}: the harmonic mean of \textit{precision} and \textit{recall}.
\end{itemize}

Because the dataset used in our experiments is imbalanced, accuracy is not the best measure to evaluate the performance of classifiers. F1-score would be a more effective measure since it considers both precision and recall measures. Table \ref{tab:evaluation} shows the performance of classifiers with/without data augmentation techniques used for DeepCapture. DeepCapture produced the best results in all metrics except precision (accuracy: 85\%, precision: 91\%, recall: 85\%, F1-score: 88\%). The existing solutions (SVM~\cite{annadatha2018image} with DA, RSVM~\cite{annadatha2018image} with DA, and CNN-SVM~\cite{shang2016image}) achieved high precision, but their recall was poor. Interestingly, traditional machine learning-based solutions (SVM~\cite{annadatha2018image} and RSVM~\cite{annadatha2018image}) failed to achieve a very low F1-score, less than 20\%, without the training samples generated by the proposed data augmentation method. In contrast with those existing techniques, deep learning-based solutions (DeepCapture and CNN-SVM~\cite{shang2016image}), achieved an F1-score of 85\% and 82\%, respectively, without data augmentation.

\begin{table}[!ht]
\centering
\caption{Performance of classifiers (DA represents ``Data Augmentation'').}
\vspace{2mm}
\label{tab:evaluation}
\begin{tabular}{|p{4.2cm}||p{1.2cm}|p{1.2cm}|p{1.2cm}|p{1.2cm}|}
\Xhline{3\arrayrulewidth}
\hline
 Model & Acc. & Pre. & Rec. & F1.\\
\hline
\textbf{DeepCapture} & \textbf{85\%} & \textbf{91\%} & \textbf{85\%} & \textbf{88\%}\\
DeepCapture without DA & 81\% & 90\% & 81\% & 85\%\\
SVM~\cite{annadatha2018image} & 51\% & 50\% & 09\% & 15\%  \\ 
SVM~\cite{annadatha2018image} with DA & 71\% & 96\% & 36\%  & 52\%  \\
RSVM~\cite{annadatha2018image} & 53\% & 52\% & 11\% & 18\%  \\
RSVM~\cite{annadatha2018image} with DA  & 73\% & 98\% & 42\% & 59\%  \\
CNN-SVM~\cite{shang2016image} & 76\% & 99\% & 71\% & 82\%  \\
CNN-SVM~\cite{shang2016image} with DA & 84\% & 90\% & 83\% & 86\%  \\
\hline
\Xhline{3\arrayrulewidth}
\end{tabular}
\end{table}

We compare DeepCapture against existing solutions (SVM~\cite{annadatha2018image}, RSVM~\cite{annadatha2018image} and CNN-SVM~\cite{shang2016image}) with respect to the training and testing times. Training time refers to the time taken to train a model with training samples. Testing time refers to the time taken to perform classification with all testing samples. Table~\ref{tab:evaluation4} shows the training and testing times of all classifiers. DeepCapture took 300.27 seconds for training and 5.79 seconds for testing. CNN-based solutions such as DeepCapture and CNN-SVM outperformed SVM and RSVM with respect to the training time. However, DeepCapture produced the worst result with respect to the training time. We surmise that the testing time of XGBoost is relatively slower than other classifiers such as SVM and RSVM because XGBoost is an ensemble of multiple regression trees. For a single image, however, the average testing time of DeepCapture was only 0.0017 seconds. Hence, we believe that the testing time of DeepCapture would be practically acceptable.

\begin{table}[!ht]
\centering
\caption{Training and testing times (sec.) of classifiers.}
\vspace{2mm}
\label{tab:evaluation4}
\begin{tabular}{|p{4.5cm}||p{2cm}|p{2cm}|}
\Xhline{3\arrayrulewidth}
\hline
 Model & Training time & Testing time\\
\hline
\textbf{DeepCapture} & \textbf{300.27} &  \textbf{5.79}\\
SVM (Annadatha et al.~\cite{annadatha2018image})  & 2000.00 & 0.01  \\
RSVM (Annadatha et al.~\cite{annadatha2018image}) & 2000.00 & 0.01  \\
CNN-SVM (Shang et al.~\cite{shang2016image}) & 320.24 &  0.03  \\
\hline
\Xhline{3\arrayrulewidth}
\end{tabular}
\end{table}

To test the robustness of classifiers against new and unseen image spam emails, we evaluate the performance of DeepCapture with cross data training. For cross data training, we trained classifiers on ham and spam images collected from one specific source, and evaluated the performance of classifiers against a different unseen dataset.

For training, we used 6,024 samples collected from ``SpamArchive spam'' and ``Normal image ham'' datasets, while for testing, we used 2,289 samples collected from ``Personal spam'' and ``Personal ham'' datasets. To make classifiers more robust against the unseen dataset, we additionally created 5,190 ham-like and 786 spam-like image samples with the images in the training set through our data augmentation techniques. Those image samples are used for training only. Table~\ref{tab:evaluation2} shows the evaluation results for the first cross data training scenario. DeepCapture achieved an F1-score of 72\% and outperformed the other classifiers. Surprisingly, F1-scores of all classifiers, including DeepCapture itself, are less than 35\% without the training samples created by data augmentation, indicating that our data augmentation techniques are necessary to process unseen and unexpected image samples. 

\begin{table}[!ht]
\centering
\caption{Performance of classifiers with a cross data training scenario (training dataset: ``SpamArchive spam'' and ``Normal image ham'' datasets; and testing dataset: ``Personal spam'' and ``Personal ham'' datasets).}
\vspace{2mm}
\label{tab:evaluation2}
\begin{tabular}{|p{4.2cm}||p{1.2cm}|p{1.2cm}|p{1.2cm}|p{1.2cm}|}
\Xhline{3\arrayrulewidth}
\hline
 Model & Acc. & Pre. & Rec. & F1.\\
\hline
\textbf{DeepCapture} & \textbf{71\%} & \textbf{81\%} & \textbf{71\%} & \textbf{72\%}\\
DeepCapture without DA & 36\% & 37\% & 34\% & 35\%\\
SVM~\cite{annadatha2018image} & 89\% & 14\% & 10\% & 12\%  \\ 
SVM~\cite{annadatha2018image} with DA & 65\% & 45\% & 22\%  & 29\%  \\
RSVM~\cite{annadatha2018image} & 90\% & 12\% & 11\% & 13\%  \\
RSVM~\cite{annadatha2018image} with DA  & 69\% & 58\% & 27\% & 30\%  \\
CNN-SVM~\cite{shang2016image} & 35\% & 35\% & 34\% & 35\%  \\
CNN-SVM~\cite{shang2016image} with DA & 68\% & 73\% & 45\% & 55\%  \\
\hline
\Xhline{3\arrayrulewidth}
\end{tabular}
\end{table}

As another cross data training scenario, we used 2,289 samples collected from ``Personal spam'' and ``Personal ham'' datasets for training while we used 6,024 samples collected from ``SpamArchive spam'' and ``Normal image ham'' datasets for testing. Again, to make classifiers more robust against the unseen dataset, we additionally created 4,497 ham-like and 5,214 spam-like image samples with the images in the training set through our data augmentation techniques. Those image samples are used for training only. Table~\ref{tab:evaluation3} shows the evaluation results for the second cross data training scenario. DeepCapture and RSVM~\cite{annadatha2018image} with DA achieved an F1-score of 76\% and outperformed the other classifiers. F1-scores of all classifiers, including DeepCapture, are less than 40\% without the training samples created by data augmentation. 

We note that in the second cross data training scenario, RSVM~\cite{annadatha2018image} with DA also produced the best classification results comparable with DeepCapture. We surmise that underlying dataset differences may explain this. In the ``Personal spam'' dataset, the ratio of spam to ham image samples is approximately 1.9:1 while in the ``SpamArchive spam'' dataset, the ratio of spam to ham image samples is approximately 6.4:1. These results demonstrate that the performance of RSVM~\cite{annadatha2018image} with DA can significantly be affected by the class distribution of samples. In contrast, DeepCapture overall works well regardless of the imbalanced class distribution of samples.


\begin{table}[!ht]
\centering
\caption{Performance of classifiers with a cross data training scenario (training dataset: “Personal spam” and “Personal ham” datasets; and testing dataset: “SpamArchive spam” and “Normal image ham” datasets).}
\vspace{2mm}
\label{tab:evaluation3}
\begin{tabular}{|p{4.2cm}||p{1.2cm}|p{1.2cm}|p{1.2cm}|p{1.2cm}|}
\Xhline{3\arrayrulewidth}
\hline
 Model & Acc. & Pre. & Rec. & F1.\\
\hline
\textbf{DeepCapture} & \textbf{73\%} & \textbf{82\%} & \textbf{72\%} & \textbf{76\%}\\
DeepCapture without DA & 31\% & 47\% & 32\% & 38\%\\
SVM~\cite{annadatha2018image} & 74\% & 61\% & 14\% & 22\%  \\ 
SVM~\cite{annadatha2018image} with DA & 60\% & 84\% & 52\%  & 64\%  \\
RSVM~\cite{annadatha2018image} & 82\% & 71\% & 22\% & 34\%  \\
RSVM~\cite{annadatha2018image} with DA  & 62\% & 94\% & 67\% & 76\%  \\
CNN-SVM~\cite{shang2016image} & 24\% & 42\% & 23\% & 30\%  \\
CNN-SVM~\cite{shang2016image} with DA & 64\% & 69\% & 47\% & 56\%  \\
\hline
\Xhline{3\arrayrulewidth}
\end{tabular}
\end{table}

\section{Related work}
\label{sec:Rw}

To avoid spam analysis and detection, spammers introduced the image spam technique to replace text spam messages with images. This strategy would be an effective technique to circumvent the text analysis of emails, which are commonly used in spam filters~\cite{Biggio11:spam}. To detect image spam emails, several classification methods have been proposed~\cite{Fumera06:spam,shang2016image,kim2017analysis,kumar2017svm,annadatha2018image,fatichah2019image}. However, the solutions offered so far exhibit several critical weaknesses. Existing detection techniques can be categorized into two approaches: (1) keyword-based analysis and (2) image classification.

\subsection{Keyword-based analysis} 

Keyword-based analysis is to extract texts from a given image and analyze them using a text-based spam filter. Several techniques~\cite{Fumera06:spam,kim2017analysis,kumar2017svm} using keyword analysis were introduced. Also, this approach was deployed in real-world spam filters such as SpamAssassin (\url{https://spamassassin.apache.org/}). Unsurprisingly, the performance of this approach depends on the performance of optical character recognition (OCR). Sophisticated spammers can intentionally embed abnormal text characters into an image, which cannot be recognized by typical OCR programs but can still be interpreted by human victims. The performance of keyword-based spam detection methods could be degraded significantly against such image spam emails. Moreover, a high processing cost of OCR is always required for analyzing images. Therefore, in this paper, we propose an image spam detection method in the direction of establishing an image classifier to distinguish spam images from ham images.


\subsection{Image classification} 

To address the high processing cost issue of keyword-based analysis, some researchers have tried to develop image spam detection methods using low-level features that are directly extracted from images. Annadatha et al.~\cite{annadatha2018image} demonstrated that image spam emails could be detected with high accuracy using either Principal Component Analysis (PCA) or Support Vector Machines (SVM). To build a classifier, they manually selected 21 features (e.g., image color, object edges) that can be extracted from spam and ham images. Shang et al.~\cite{shang2016image} proposed an alternative image classification method using a CNN model and an SVM classifier together, which is composed of 13 layers. The CNN model proceeds classification in the last fully connected layer. However, they use the output from the last fully connected layer as the input for the SVM classifier. In this paper, we develop a more compact CNN-XGBoost model consisting of 8 layers. Our evaluation results show that DeepCapture outperforms Shang et al.'s architecture in terms of detection accuracy. Fatichah et al.~\cite{fatichah2019image} also discussed the possibility of CNN models to detect image spam. Unlike other previous studies, they focused on building CNN models to detect the image spam on Instagram (\url{https://www.instagram.com/}), a social photo-sharing service. They evaluated the performance of four pre-trained CNN models (3-layer, 5-layer, AlexNet, and VGG16) with 8,000 images collected from Instagram. They found that the VGG16 architecture achieves the best accuracy (about 0.84) compared with the other models. Since VGG16 is a pre-trained network and its performance is not advantageous, we do not directly compare DeepCapture with VGG16.


We note that the performances of previous methods have been evaluated on different data sets with different configurations. Therefore, we cannot directly compare their reported performances. In this paper, we needed to reimplement their models and used the publicly available datasets to compare the performance of DeepCature with those of the best existing models (SVM~\cite{annadatha2018image}, RSVM~\cite{annadatha2018image} and CNN-SVM~\cite{shang2016image}).


\section{Conclusion}
\label{sec:CS}

In this paper, we proposed a new image spam email detection tool called DeepCapture. To overcome the performance degrade of existing models against entirely new and unseen datasets, we developed a classifier using CNN-XGBoost and data augmentation techniques tailored towards the image spam detection task. To show the feasibility of DeepCapture, we evaluate its performance with three publicly available datasets consisting of spam and non-spam image samples. The experimental results demonstrated that DeepCapture is capable of achieving 88\% F1-score, which has 6\% improvement over the best existing spam detection model, CNN-SVM~\cite{shang2016image}, with an F1-score of 82\%. Furthermore, DeepCapture outperforms other classifiers in cross data training scenarios to evaluate the performance of classifiers with the new and unseen dataset.

For future work, we plan to develop more sophisticated data augmentation methods to add a more real-like synthetic dataset effectively. In addition, we will increase the size of the dataset and examine any changes in detection accuracy. It would also be interesting to add the functionality of DeepCapture to an open-source project such as SpamAssassin.

\section*{Acknowledgement} 
\label{sec:ack}
\begin{small}
Hyoungshick Kim is the corresponding author. This work has been supported in part by the Cyber Security Research Centre Limited whose activities are partially funded by the Australian Government's Cooperative Research Centres Programme and the NRF grant (No. 2017H1D8A2031628) and the ITRC Support Program (IITP-2019-
2015-0-00403) funded by the Korea government. The authors would like to thank all the anonymous reviewers for their valuable feedback.
\end{small}

\bibliographystyle{splncs04}
\bibliography{mybibliography}

\end{document}